\documentclass[letterpaper, 10pt, journal, twoside]{IEEEtran}

\IEEEoverridecommandlockouts    % This command is only needed if
\usepackage{graphics}           
\usepackage{times}              
\usepackage{amsmath}            
\usepackage{amssymb}            
\usepackage{graphicx}
\usepackage{algorithm}
\usepackage[noend]{algpseudocode}
\usepackage{booktabs}
\usepackage{color}
\usepackage{listings}
\usepackage{subfiles}
\usepackage{hyperref}
\usepackage[nocompress]{cite} % sorts citations automatically, e.g., "[1],[2],[3]" instead of "[2],[3],[1]".
\definecolor{instructioncolor}{rgb}{.5,.5,.5}

\usepackage[font=small]{caption}

\def\eqref#1{Eq.~(\ref{#1})}

\makeatletter
\usepackage{xspace}
\DeclareRobustCommand\onedot{\futurelet\@let@token\@onedot}
\def\@onedot{\ifx\@let@token.\else.\null\fi\xspace}

\makeatother

\usepackage{array}
\newcolumntype{L}[1]{>{\raggedright\let\newline\\\arraybackslash\hspace{0pt}}m{#1}}
\newcolumntype{C}[1]{>{\centering\let\newline\\\arraybackslash\hspace{0pt}}m{#1}}
\newcolumntype{R}[1]{>{\raggedleft\let\newline\\\arraybackslash\hspace{0pt}}m{#1}}

\def\argmax{\mathop{\rm argmax}}

\usepackage{amssymb}
\usepackage{amsmath}
\usepackage{subcaption}
\usepackage{stfloats}
\usepackage{verbatim}
\usepackage{graphicx}
\usepackage{cite}
\usepackage{booktabs}
\usepackage{multirow}
\usepackage{caption}
\usepackage{array}
\usepackage{wrapfig}

\usepackage[nameinlink]{cleveref}

\Crefname{section}{Sec.}{Sec.}
\Crefname{figure}{Fig.}{Fig.}
\Crefname{equation}{Eq.}{Eq.}
\Crefname{table}{Tab.}{Tab.}

\usepackage[printonlyused,withpage,nolist,nohyperlinks]{acronym}
\begin{acronym}

\acro{2D}{two-dimensional}

\acro{3D}{three-dimensional}

\acro{AHRS}{attitude and heading reference system}

\acro{AUV}{autonomous underwater vehicle}

\acro{CPP}{Chinese Postman Problem}

\acro{DoF}{degree of freedom}
\acrodefplural{DoF}[DoFs]{degrees of freedom}

\acro{DVL}{Doppler velocity log}

\acro{FSM}{finite state machine}

\acro{IMU}{inertial measurement unit}

\acro{LBL}{Long Baseline}

\acro{MCM}{mine countermeasures}

\acro{MDP}{Markov decision process}
\acrodefplural{MDP}[MDPs]{Markov decision processes}

\acro{POMDP}{Partially Observable Markov Decision Process}
\acrodefplural{POMDP}[POMDPs]{Partially Observable Markov Decision Processes}

\acro{PRM}{Probabilistic Roadmap}
\acrodefplural{PRM}[PRM]{Probabilistic Roadmaps}

\acro{ROI}{region of interest}
\acrodefplural{ROI}[ROIs]{regions of interest}

\acro{ROS}{Robot Operating System}

\acro{ROV}{remotely operated vehicle}

\acro{RRT}{Rapidly-exploring Random Tree}
\acrodefplural{RRT}[RRTs]{Rapidly-exploring Random Trees}

\acro{SLAM}{Simultaneous Localisation and Mapping}

\acro{SSE}{sum of squared errors}

\acro{STOMP}{Stochastic Trajectory Optimization for Motion Planning}

\acro{TRN}{Terrain-Relative Navigation}

\acro{UAV}{unmanned aerial vehicle}
\acro{UGV}{unmanned ground vehicle}

\acro{USBL}{Ultra-Short Baseline}

\acro{IPP}{informative path planning}

\acro{FoV}{field of view}
\acrodefplural{FoV}[FoVs]{fields of view}

\acro{CDF}{cumulative distribution function}

\acro{ML}{maximum likelihood}

\acro{RMSE}{root mean squared error}
\acro{MLL}{mean log loss}

\acro{GP}{Gaussian Process}
\acrodefplural{GP}[GPs]{Gaussian Processes}

\acro{KF}{Kalman Filter}

\acro{IP}{Interior Point}
\acro{BO}{Bayesian Optimization}

\acro{SE}{squared exponential}

\acro{UI}{uncertain input}

\acro{MCL}{Monte Carlo Localisation}
\acro{AMCL}{Adaptive Monte Carlo Localisation}

\acro{SSIM}{Structural Similarity Index}

\acro{MAE}{mean absolute error}
\acro{AUSE}{Area Under the Sparsification Error curve}

\acro{AL}{active learning}
\acro{DL}{deep learning}
\acro{CNN}{convolutional neural network}
\acro{GNN}{graph neural network}
\acro{MLP}{multilayer perceptron}

\acro{MC}{Monte-Carlo}

\acro{GSD}{ground sample distance}

\acro{BALD}{Bayesian active learning by disagreement}

\acro{fCNN}{fully convolutional neural network}
\acro{FCN}{fully convolutional neural network}

\acro{CMA-ES}{covariance matrix adaptation evolution strategy}

\acro{FoV}{field of view}

\acro{mIoU}{mean Intersection-over-Union}
\acro{ECE}{expected calibration error}
\acro{II}{information integral}

\acro{GAN}{Generative Adversarial Network}
\acro{POMCP}{Partially Observable Monte-Carlo Planning}
\acro{MCTS}{Monte-Carlo tree search}

\acro{RL}{reinforcement learning}
\acro{PPO}{proximal policy optimisation}
\acro{SAC}{Soft Actor-Critic}
\acro{DQN}{Deep Q-Network}

\end{acronym}

\title{Towards Map-Agnostic Policies for Adaptive Informative Path Planning}

\author{\hspace{5mm}Julius R\"{u}ckin$^{1}$ \hfill David Morilla-Cabello$^{2}$ \hfill Cyrill Stachniss$^{1,4}$ \hfill Eduardo Montijano$^{2}$ \hfill Marija Popovi\'{c}$^{3}$\hspace{5mm}
\thanks{Manuscript received: Oct, 21, 2024; Revised Jan, 24, 2025; Accepted Mar, 18, 2025. This paper was recommended for publication by Editor Giuseppe Loianno upon evaluation of the Associate Editor and Reviewers’ comments.}
\thanks{$^{1}$J. R\"{u}ckin. and C. Stachniss are with the Center for Robotics, University of Bonn, Germany. $^{2}$D. Morilla-Cabello and E. Montijano are with DIIS-I3A, Universidad de Zaragoza, Spain. $^{3}$M. Popovi\'{c} is with the MAVLab, TU Delft, Netherlands. $^{4}$C. Stachniss is also with the Lamarr Institute for Machine Learning and Artificial Intelligence, Germany.}
\thanks{This work has been funded by the Deutsche Forschungsgemeinschaft (DFG, German Research Foundation) under Germany's Excellence Strategy, EXC-2070 -- 390732324 (PhenoRob), DGA project T45\_23R, MCIN/AEI/ERDF/European Union NextGenerationEU/PRTR project PID2021-125514NB-I00, ONR grant N62909-24-1-2081 and grant FPU20-06563.
Corresponding author: \texttt{jrueckin@uni-bonn.de}.}
\thanks{Digital Object Identifier (DOI): see top of this page.}
}
\begin{document}
\maketitle

%%%%%%%%%%%%%%%%%%%%%%%%%%%%%%%%%%%%%%%%%%%%%%%%%%%%%%%%%%%%%%%%%%%%%%%%%%%%%%%%
\begin{abstract}
Robots are frequently tasked to gather relevant sensor data in unknown terrains. A key challenge for classical path planning algorithms used for autonomous information gathering is adaptively replanning paths online as the terrain is explored given limited onboard compute resources. Recently, learning-based approaches emerged that train planning policies offline and enable computationally efficient online replanning performing policy inference. These approaches are designed and trained for terrain monitoring missions assuming a single specific map representation, which limits their applicability to different terrains. To address this limitation, we propose a novel formulation of the adaptive \acl{IPP} problem unified across different map representations, enabling training and deploying planning policies in a larger variety of monitoring missions. Experimental results validate that our novel formulation easily integrates with classical non-learning-based planning approaches while maintaining their performance. Our trained planning policy performs similarly to state-of-the-art map-specifically trained policies. We validate our learned policy on unseen real-world terrain datasets.
\end{abstract}

\begin{IEEEkeywords}
Motion and Path Planning,  Aerial Systems: Perception and Autonomy, Reinforcement Learning
\end{IEEEkeywords}

%===============================================================================

\section{Introduction}
\label{sec:intro}
	
\IEEEPARstart{D}{ecision-making} under uncertainty in unknown terrains is a crucial skill for autonomous robots in many real-world scenarios, such as exploration~\cite{liu2023learning, lluvia2021active}, environmental monitoring~\cite{rayas2022informative, marchant2014icra, viseras2021wildfire}, precision agriculture~\cite{popovic2020informative, vivaldini2019uav}, and search and rescue~\cite{niroui2019deep, singh2009nonmyopic}. To complete their mission goals, robots gather relevant information about the terrain using onboard sensors. A key challenge is to adapt planned paths online based on newly incoming noisy measurements under limited onboard compute and mission budget as the robot's understanding of the terrain evolves. This problem is known in the literature as the adaptive \acf{IPP} problem~\cite{choudhury2020adaptive, hitz2017adaptive, marchant2014icra, hollinger2014sampling, ott2023sboaippms, popovic2024learning}.

Specifically, this work examines the problem of mapping user-defined areas of interest using a budget-constrained robot with noisy onboard sensors~\cite{marchant2014icra, hitz2017adaptive, popovic2020informative, hollinger2014sampling}. To this end, the robot adaptively replans paths online to maximise the information gathered about the initially unknown areas of interest based on the evolving understanding of the terrain. The gathered information is captured in a continuously updated terrain map using newly acquired sensor measurements.

\begin{figure}[!t]
    \centering
    \includegraphics[width=\columnwidth]{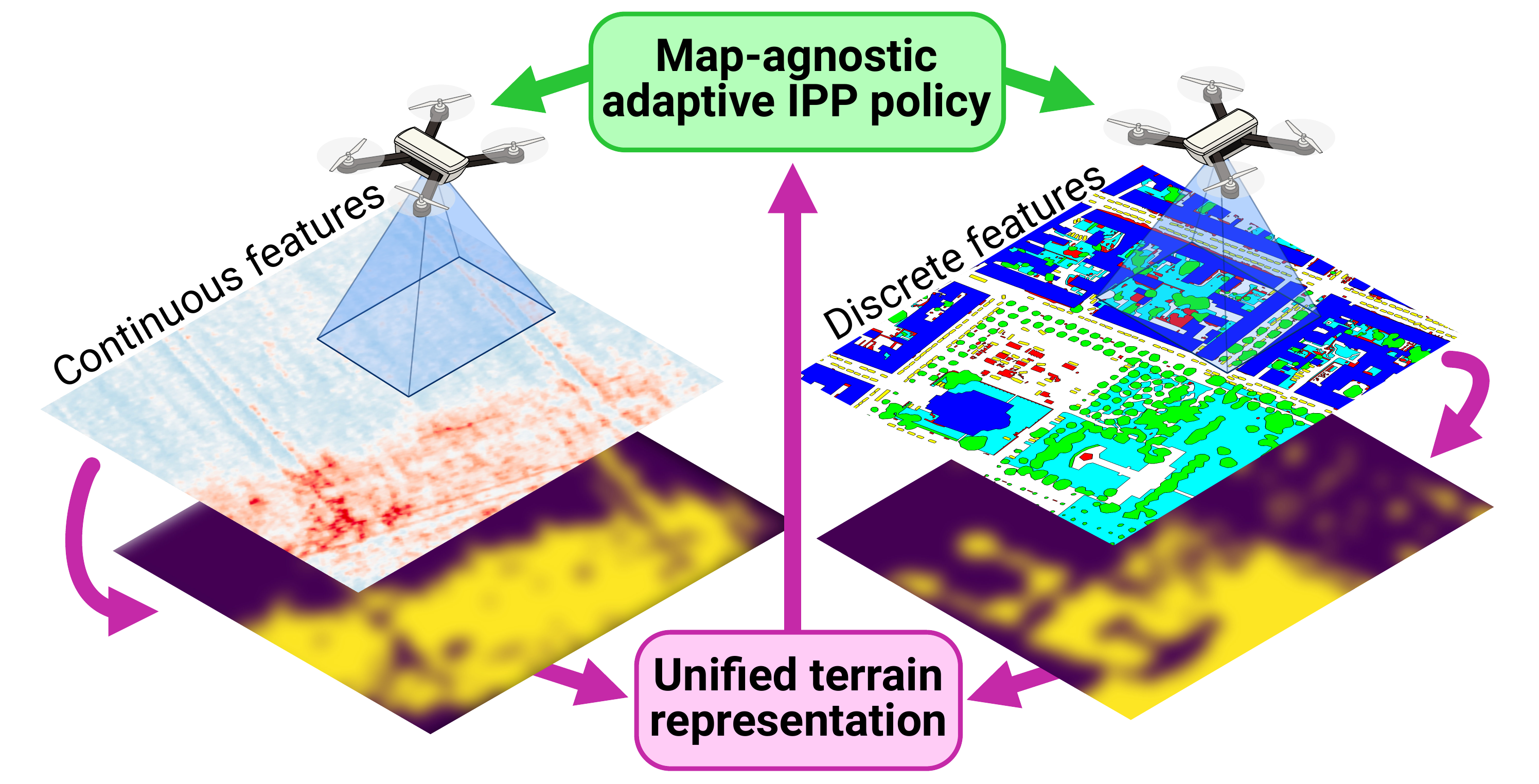}
    \caption{Robots perform continuous- or discrete-valued terrain feature monitoring missions, e.g. mapping surface temperature or urban semantics. We transform mission-specific terrain map representations, e.g. Gaussian processes or occupancy grid maps, into a novel unified state representation for adaptive \acf{IPP}. In this way, we design and train a single map-agnostic planning policy applicable to largely varying terrain monitoring missions.}
    \label{F:approach_overview}
\end{figure}

Various adaptive \ac{IPP} approaches emerged for different to-be-mapped information, i.e. terrain features, of interest during a mission. Mapping continuous-valued terrain features, e.g. bacteria levels~\cite{hitz2017adaptive} or signal strength~\cite{marchant2014icra}, is commonly performed using Gaussian processes~\cite{hitz2017adaptive, ott2024approximate, vivaldini2019uav} or Kalman filter~\cite{popovic2020informative, ruckin2022adaptive} map representations. Mapping discrete-valued terrain features, e.g. crop-weed~\cite{westheider2023multi} or rural area semantic segmentation~\cite{popovic2020informative}, is commonly performed using grid maps. Based on the current map, non-learning-based planning algorithms iteratively select candidate paths and evaluate their expected information value~\cite{hitz2017adaptive, marchant2014icra, hollinger2014sampling, ott2023sboaippms}. These approaches can be adapted for different map representations. However, they tend to be too compute-intensive for frequent online replanning as they rely on costly evaluations of many potential future paths. To overcome these issues, learning-based approaches have been proposed. These methods train adaptive \ac{IPP} policies offline in simulation and perform compute-efficient policy inference at deployment~\cite{niroui2019deep, liu2023learning, cao2023adriadne, choi2021adaptive, wei2020informative, cao2023catnipp, ruckin2022adaptive}. Although learning-based approaches show promising performance, they are specifically designed for and trained on a single terrain map representation. This prohibits their direct application to a larger variety of terrain monitoring missions. 

We argue that the broad pool of existing adaptive \ac{IPP} approaches should be viewed along two dimensions: the map-specific formulation modelling the adaptive \ac{IPP} problem and the algorithm used to offline-train or online-search the planning policy. The formulation of the adaptive \ac{IPP} problem is the most critical design decision to ensure the unified applicability of planning policies across various terrain monitoring missions. This motivates the need for a map-agnostic formulation of the adaptive \ac{IPP} terrain monitoring problem that directly integrates with any (non)-learning-based policy search algorithm used for adaptive \ac{IPP}. Particularly, this formulation ensures training and deploying learned policies in largely varying terrain monitoring missions.     

The main contribution of this paper is such a novel map-agnostic formulation of the adaptive \ac{IPP} problem for terrain monitoring illustrated in \Cref{F:approach_overview}. Our formulation unifies continuous-valued, i.e. regression, and discrete-valued, i.e. classification, terrain feature monitoring for adaptive \ac{IPP} policies. To achieve this, we unify state space representations across terrain map representations utilised for replanning online. Based on this unified state space and a new reward function, we train and deploy a single generally applicable planning policy on previously unmet variations of terrain monitoring missions using \ac{RL}. %To this end, we condition the learned planning policy on the user-defined and map-specific mission hyperparameters to increase performance across diverse monitoring missions.

In sum, we make the following claims. First, our map-agnostic planning policy trained and deployed on vastly varying simulated terrain monitoring missions performs on par or better than state-of-the-art map-specifically trained policies and non-learning-based adaptive \ac{IPP} approaches. Second, our map-agnostic policy performs similarly to state-of-the-art adaptive \ac{IPP} approaches on various real-world terrain datasets. Third,  in our experiments, we demonstrate that our map-agnostic adaptive \ac{IPP} formulation easily integrates with previous non-learning-based state-of-the-art adaptive \ac{IPP} algorithms while maintaining or improving their performance. We will open-source our code for usage by the community at: \url{https://github.com/dmar-bonn/ipp-rl-gen}.

%===============================================================================

\section{Related Work} %Work in progress
\label{sec:related_work}

% Goal of the work, rough problem statement, non-adaptive vs. adaptive approaches.
This work addresses the problem of robotic information gathering in initially unknown terrains where certain areas are considered more interesting than others, e.g. temperature hotspots~\cite{popovic2020informative, choi2021adaptive} or search and rescue victims~\cite{niroui2019deep, singh2009nonmyopic}. This problem is known as the adaptive \ac{IPP} problem~\cite{hitz2017adaptive, marchant2014icra, hollinger2014sampling, ott2023sboaippms, popovic2024learning}, where the aim is to efficiently discover and precisely map the areas of interest using a resource-constrained robot, e.g. an \ac{UAV} with limited battery capacity~\cite{popovic2020informative, vivaldini2019uav, viseras2021wildfire}. Various adaptive \ac{IPP} approaches have been proposed, which actively replan paths online during a mission based on the robot's state and previously collected measurements. In contrast, often less efficient non-adaptive approaches, e.g. coverage paths~\cite{galceran2013survey, tan2021comprehensive}, pre-compute static paths that cannot be modified during a mission.

% Non-learning-based adaptive approaches (sampling-based, optimisation-based, geometry-based)
Methods for adaptive \ac{IPP} can be categorised into non-learning-based and learning-based planning approaches. Non-learning-based approaches have been successfully applied to many different variants of the adaptive \ac{IPP} problem, such as exploration~\cite{lluvia2021active, liu2023learning}, search and rescue~\cite{choudhury2020adaptive, singh2009nonmyopic}, and terrain monitoring~\cite{hitz2017adaptive, vivaldini2019uav, hollinger2014sampling}. Sampling-based methods solve the adaptive \ac{IPP} problem by iteratively (re-)sampling potential paths and evaluating their information value based on the robot's terrain understanding, building upon established sampling-based search, such as receding horizon planning~\cite{hollinger2014sampling} or \ac{MCTS}~\cite{choudhury2020adaptive, ott2023sboaippms}. Optimisation-based methods utilise derivative-free optimisation, such as evolutionary algorithms~\cite{popovic2020informative, hitz2017adaptive} or Bayesian optimisation~\cite{vivaldini2019uav, marchant2014icra}, directly maximising the information acquired along the path. 
%Geometry-based methods collect potential next robot positions at frontiers of explored and unexplored terrain space, evaluate and select the position with maximal information value~\cite{yamauchi1997frontier, saroya2020online}. 
Although non-learning-based planning methods for adaptive \ac{IPP} show promising results, they tend to be computationally inefficient~\cite{popovic2020informative, cao2023catnipp, ruckin2022adaptive} as they evaluate expensive-to-compute information criteria for many potential future paths, prohibiting fast online replanning or sacrificing path quality. Further, these approaches directly use mission-specific terrain map representations to design the planning state space, which requires adapting planning methods for deploying them in monitoring missions with different terrain maps.

% Learning-based approaches (imitation learning vs. reinforcement learning)
Recently, learning-based methods were proposed to tackle the adaptive \ac{IPP} problem, providing higher compute efficiency and achieving similar or better planning performance. This is done by shifting the computational burden to an offline training phase, simulating many terrain monitoring missions, and inferring the learned planning policy at deployment~\cite{chen2021zeroshot, wei2020informative, cao2023catnipp, ruckin2022adaptive, westheider2023multi, choi2021adaptive, liu2023learning, viseras2021wildfire, vashisth2024deep}. \ac{RL} methods have been proposed for specific adaptive \ac{IPP} applications, such as terrain exploration~\cite{niroui2019deep, liu2023learning, cao2023adriadne} and monitoring~\cite{choi2021adaptive, wei2020informative, cao2023catnipp, ruckin2022adaptive, vashisth2024deep}. These works mainly differ in their reward function design influenced by the mission goal and terrain map representation, and the used policy networks trained with different \ac{RL} algorithms. Methods for exploration design reward functions measuring coverage of the terrain~\cite{liu2023learning, cao2023adriadne}, while works considering efficiently finding and precisely mapping areas of interest commonly reward decreasing terrain map uncertainty in these areas~\cite{viseras2021wildfire, cao2023catnipp, ruckin2022adaptive, westheider2023multi}. All these approaches maintain a mission-specific spatial terrain map representation to fuse onboard measurements, such as occupancy maps~\cite{niroui2019deep, liu2023learning, yang2023learning, westheider2023multi, cao2023adriadne} or sub-sampled Gaussian processes~\cite{choi2021adaptive, wei2020informative, cao2023catnipp, vashisth2024deep}. These approaches directly use mission-specific terrain map representations to design the planning algorithm's map-specific state space and train the adaptive \ac{IPP} policy on this map-specific state space.

% Problems with current RL approaches and how our work solves them (partially)
Overall, all previous adaptive \ac{IPP} approaches propose terrain map-specific solutions, assuming either occupancy maps~\cite{niroui2019deep, popovic2020informative, liu2023learning, yang2023learning, westheider2023multi, cao2023adriadne}, pre-trained Gaussian processes~\cite{hitz2017adaptive, wei2020informative, choi2021adaptive, cao2023catnipp, ott2023sboaippms, vashisth2024deep} or Kalman filters~\cite{popovic2020informative, ruckin2022adaptive}, as their planning state representation. Thus, these methods require adaptation and re-training as the map, and hence, planning state representations change. This prohibits the application of learned policies to various monitoring missions that require different map representations. In contrast, we present a novel map-agnostic state formulation for adaptive \ac{IPP} unifying terrain monitoring missions across various map representations. Combining this map-agnostic state space with a new reward function, we train a single planning policy using \ac{RL} that is applicable to continuous- and discrete-valued terrain feature monitoring missions. 
%We introduce a set of hyperparameters for adaptive \ac{IPP} problems to condition the policy on user-defined characteristics potentially varying across missions, e.g. interesting terrain features.
	
%===============================================================================

\section{Problem Formulation}
\label{sec:problem_formulation}

This work aims to formulate the adaptive \ac{IPP} problem for terrain monitoring~\cite{meliou2007nonmyopic, hollinger2013active, marchant2014icra, hollinger2014sampling, popovic2020informative} in a terrain map-agnostic fashion to offline-learn or online-solve planning policies across different monitoring missions and map representations without adapting or re-training policies. 
% Variable and setup definition
We consider a robot with pose $\mathbf{p}_t \in \mathbb{R}^{D_r}$  at time $t,$ moving in an \textit{a priori} unknown terrain. 
% Terrain definition
The terrain $\xi \subset \mathbb{R}^{D_e}$ is characterised by its initially unknown and stationary feature field $F: \xi \to \mathcal{F},$ where $\mathcal{F}$ is the mission-specific continuous or discrete terrain feature space. The goal is to estimate and precisely map the terrain feature field $F$ in interesting areas,
\begin{equation} \label{eq:roi}
     \xi_I = \{\mathbf{x} \in \xi \mid F(\mathbf{x}) \in \mathcal{F}_{I}\} \subseteq \xi \, ,
\end{equation}
where $\mathcal{F}_{I} \subseteq \mathcal{F}$ is the user-defined subset of feature values qualifying the interest in a point $\mathbf{x} \in \xi$, e.g. a value range or semantic classes, with each $\mathbf{x} \in \xi_I$ being of equal interest.

% Platform definition and "dynamics"
To accomplish this objective, the robot is equipped with a sensor to collect measurements $z \in \mathcal{Z}$ from the terrain, e.g. semantically segmented RGB images, thermal images, or radiation levels. At each time step $t$, the measurements provide noisy information about $F$ according to $z_t \sim p(z \,|\, \mathbf{p}_t, F)$ and are used to model a stochastic process $\hat{F}_t$ over all possible terrain feature field functions $F$,
\begin{equation} \label{eq:map_belief}
    \hat{F}_t \sim p(F \mid z_{1:t}, \mathbf{p}_{1:t},\theta_F)\,,
\end{equation}
where $z_{1:t}$ is the set of all collected measurements at robot poses $\mathbf{p}_{1:t}$, and $\theta_F$ indicates the chosen map representation and its hyperparameters. Most works update the belief for continuous-valued feature spaces $\mathcal{F} \subseteq \mathbb{R}$ with pre-trained Gaussian processes or Kalman filters, and for discrete-valued feature spaces $\mathcal{F} \subseteq \mathbb{N}$ with occupancy grid mapping.

%During a mission, prior adaptive \ac{IPP} methods~\cite{hitz2017adaptive, popovic2020informative, cao2023catnipp, ruckin2022adaptive, westheider2023multi} estimate unknown areas of interest $\xi_I$ by
%
%\begin{equation} \label{eq:roi_belief}
%    \hat{\xi}_{I, t} = \left\{\mathbf{x} \in \xi ~\bigl\vert~ p\big(F(\mathbf{x}) \in \mathcal{F}_{I} \mid \hat{F}_t\big) \geq c_{th}\right\} \, ,
%\end{equation}
%
%where $c_{th} \in [0,1]$ is a hand-tuned threshold at which a point $\mathbf{x} \in %\xi$ is considered to be within the true area of interest $\xi_I$ with sufficient %confidence, and confidence intervals are re-computed based on the specific map %representation $\hat{F}_t$~\cite{popovic2020informative}.

% IPP problem
We aim to find an optimal action sequence $\psi^* = (\mathbf{a}_1, \,\ldots, \mathbf{a}_N)$, where $\mathbf{a}_t \in \mathcal{A} \subseteq \mathbb{R}^{D_a}$ are relative pose changes. The action sequence $\psi^*$ maximises an information criterion $I: \mathcal A^N \times\, \xi_I \to \mathbb R$, where $\mathcal{A}^N$ encompasses all action sequences of variable length $N$, associating the sensor measurements collected while executing an action sequence $\psi$ with their information value about areas of interest $\xi_I$,
\begin{equation}
    \label{eq:aipp_problem}
    \psi^{*} = \argmax_{\psi \in \mathcal{A}^N} \,I(\psi, \xi_I), \,\,\text{s.t. } C(\psi) \leq B \,,
\end{equation}
where $C: \mathcal{A}^N \to \mathbb{R}$ is the action sequence execution cost, e.g. battery capacity or travel time, and $B \geq 0$ is the robot's fixed maximum mission budget. 
% Adaptive replanning
As $F$ and thus $\xi_I$ are \textit{a priori} unknown, \Cref{eq:aipp_problem} cannot be solved offline. The optimal action sequence $\psi^*$ in~\Cref{eq:aipp_problem} changes as $\hat{F}_t$ is updated based on new measurements. Therefore, online replanning is required to find an optimal $\psi^*$ that \textit{adaptively} focuses on areas of interest $\xi_I$ as they are discovered.

The concrete formulation of \Cref{eq:aipp_problem} depends on the specific terrain monitoring mission. Depending on the mission characteristics, the spatially mapped terrain feature space $\mathcal{F}$ might be discrete, such as semantic classes, or continuous, such as surface temperature. For a given mission, the user defines interesting features $\mathcal{F}_I \subseteq \mathcal{F}$ and chooses the map representation $\hat{F}_t$ with map hyperparameters $\theta_F$. We denote $\mathcal{H} = \{\mathcal{F}, \mathcal{F}_I, \theta_F\}$ as the set of mission hyperparameters defining the specific instantiation of \Cref{eq:aipp_problem}.

% IPP as an RL problem
As shown in previous works~\cite{niroui2019deep, liu2023learning, wei2020informative, cao2023catnipp, ruckin2022adaptive, cao2023adriadne, yang2023learning}, the adaptive \ac{IPP} problem in \Cref{eq:aipp_problem} can be transformed into an \ac{RL} problem for many terrain monitoring mission variants by
\begin{equation}\label{eq:ipp_problem_as_rl}
    \begin{aligned} 
    \pi^* &= \argmax_{\pi \in \Pi} I\left((\pi(s_1), \ldots, \pi(s_N)), \xi_I\right) \\
    &= \argmax_{\pi \in \Pi} \sum_{t=1}^{N} \gamma^{t-1} R\left(s_t, \pi(s_t), s_{t+1}, \xi_I\right),
\end{aligned}
\end{equation}
where $\pi: \mathcal{S} \to \mathcal{A}$ is a planning policy mapping state $s_t \in \mathcal{S}$ at a time step $t$ to an action $\mathbf{a}_t = \pi(s_t)$, and $\Pi$ is the function space of all possible policies. Thus, the action sequence~$\psi$ is given by $\psi = (\pi(s_1), \ldots, \pi(s_N))$. A mission- and map-specific reward function $R(s_t, \pi(s_t), s_{t+1}, \xi_I)~\in~\mathbb{R}$ resembles the information criterion $I$, rewarding taking actions $\mathbf{a}_t$ in state $s_t$ that lead to a next state $s_{t+1}$ with increased information about interesting areas $\xi_I$, and $\gamma \in [0, 1]$ is a discount factor. As interesting areas $\xi_I$ are unknown during a mission, prior adaptive \ac{IPP} methods~\cite{hitz2017adaptive, popovic2020informative, cao2023catnipp, ruckin2022adaptive, westheider2023multi} approximate unknown areas of interest $\xi_I$ using map-specifically computed confidence intervals based on hand-tuned confidence thresholds, rewarding uncertainty reduction over map belief $\hat{F}_t$ in these approximated interesting areas.

% Our work's goal
Different from existing adaptive \ac{IPP} approaches that consider terrain map-specific planning state formulations $s_t$ with approximated areas of interest, we formulate the problem in \Cref{eq:ipp_problem_as_rl} in a fully probabilistic map-agnostic fashion. To this end, we propose a planning state $s_t$ that unifies the adaptive \ac{IPP} problem across different map representations $\hat{F}_t$, allowing us to apply a single learned policy $\pi^*$ to varying terrain monitoring missions. Based on this planning state, we introduce a new reward function for \Cref{eq:ipp_problem_as_rl} enabling training or online-solving policy $\pi^*$ for different monitoring missions. 
%We use the mission hyperparameters $\mathcal{H}$ to condition learned policy $\pi^*$ on the user-defined mission characteristics $\mathcal{F}_I, \theta_F$.

%===============================================================================

\section{Our Approach}
\label{sec:approach}

\begin{figure*}[!t]
    \captionsetup[subfigure]{labelformat=empty}
    \centering
    \includegraphics[width=\textwidth]{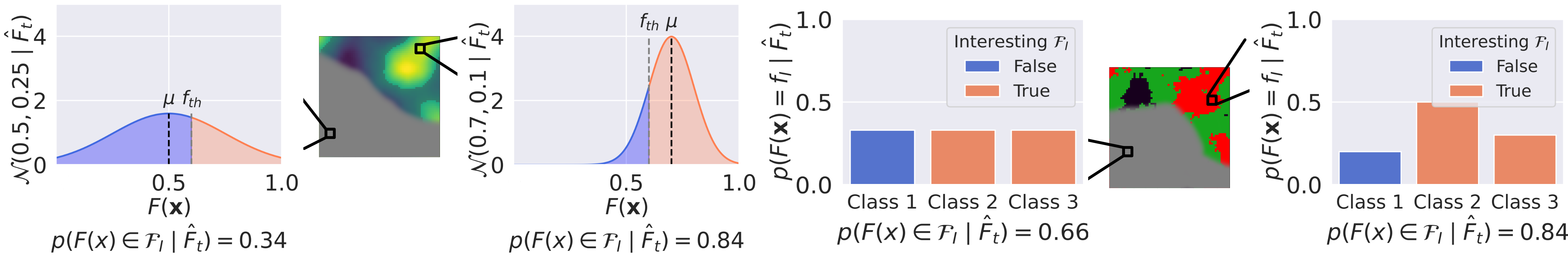}
    \caption{Our unified belief $p\big(F(\mathbf{x}) \in \mathcal{F}_{I} \mid \hat{F}_t\big)$ over interesting areas $\mathbf{x} \in \xi_I$ for continuous- (left) and discrete-valued (right) terrain features. Grey areas are unknown with large map uncertainty. (Left) Posterior normal distributions inferred from a Gaussian process or Kalman filter map representation with an interesting value threshold $f_{th} = 0.6$. The unified belief is computed by the orange area under the curve, which is larger for known interesting areas than for unknown uncertain areas. (Right) The unified belief is given by the sum of posterior probability masses over interesting classes (orange) extracted from an occupancy map representation.}
    \label{F:unified_state_space}
\end{figure*}

Our approach is conceptually depicted in \Cref{F:approach_overview}. We unify the adaptive \ac{IPP} problem formulation introduced in \Cref{sec:problem_formulation} across different map representations required to spatially capture various continuous- and discrete-valued terrain features. To this end, we view any terrain monitoring mission as a binary classification task, probabilistically splitting the terrain into unknown interesting areas $\xi_I$ (\Cref{eq:roi}) and uninteresting areas $\xi - \xi_I$. Based on this belief over interesting areas $\xi_I$, we propose a map-agnostic planning state space (\Cref{subsec:unified_state_space}) and introduce a reward function to online-solve or offline-train a planning policy across terrain monitoring missions with different map representations (\Cref{subsec:reward_function}). Last, we show how we use our state formulation and reward function to offline-train adaptive \ac{IPP} policies on varying terrain monitoring missions in simulation (\Cref{subsec:policy_training}).

\subsection{Unified Planning State Space for Adaptive IPP}
\label{subsec:unified_state_space} 

Our formulation of planning states $s_t \in \mathcal{S}$ encodes all information required to solve the adaptive \ac{IPP} problem in \Cref{eq:ipp_problem_as_rl}, i.e. the robot's state estimation, its current understanding of the terrain, and mission hyperparameters. We propose a unified belief over interesting terrain areas reusable as input to the planning policy $\pi(s_t)$ for feature fields $F$ with continuous- and discrete-valued terrain features $\mathcal{F}$ that might require different map representations $\hat{F}_t$. Assume that $\mathcal{X}_t$ is a set of points $\mathbf{x}_t \in \xi$ sampled from the terrain $\xi$ at time step $t$ at which we aim to infer the state $s_t(\mathbf{x}_t)$. Then, for each $\mathbf{x}_t \in \mathcal{X}_t$, $s_t(\mathbf{x}_t)$ is defined as 
\begin{equation} \label{eq:env_state}
    \resizebox{0.90\linewidth}{!}{
    $s_t(\mathbf{x}_t) = \left(p\big(F(\mathbf{x}_t) \in \mathcal{F}_{I} \mid \hat{F}_t\big), H\big(\hat{F}_t(\mathbf{x}_t)\big), \mathbf{p}_t, B_t, \mathcal{H}\right),$
    }
\end{equation}
where $p\big(F(\mathbf{x}_t) \in \mathcal{F}_{I} \mid \hat{F}_t\big)$ is the probability of $\mathbf{x}_t$ being part of an interesting area $\xi_I$, $H\big(\hat{F}_t(\mathbf{x}_t)\big)$ is the uncertainty of the mission-specific map belief $\hat{F}_t$ at $\mathbf{x}_t$, $\mathbf{p}_t$ is the robot's current position, $B_t \leq B$ is the robot's remaining budget, and $\mathcal{H}$ are the mission hyperparameters specifying \Cref{eq:aipp_problem}. For occupancy maps, $H\big(\hat{F}_t(\mathbf{x}_t)\big)$ is the Shannon entropy at $\mathbf{x}_t$. For Gaussian processes or Kalman filters, $H\big(\hat{F}_t(\mathbf{x}_t)\big)$ is the variance at $\mathbf{x}_t$. Our state can be integrated with any representation $\mathcal{X}_t$ of terrain $\xi$ to compute the state representation $s_t$ over $\mathcal{X}_t$. For example, it supports equidistant grids~\cite{popovic2020informative, ott2023sboaippms, westheider2023multi, ruckin2022adaptive} or randomly sampled graphs~\cite{cao2023catnipp, cao2023adriadne, vashisth2024deep, wei2020informative}.

In contrast to previous works relying on map-specific formulations of $s_t$ with binary approximations of interesting areas, our planning state formulation in \Cref{eq:env_state} introduces a fully probabilistic map-agnostic belief over interesting areas. Next, we show how to compute this map-agnostic belief $\hat{F}_{I, t} \sim p\big(F(\mathbf{x}_t) \in \mathcal{F}_{I} \mid \hat{F}_t\big)$ for continuous- and discrete-valued terrain feature mapping missions with different fully probabilistic map representations $\hat{F}_t$ as illustrated in \Cref{F:unified_state_space}. 

Consider \textit{discrete feature spaces} $\mathcal{F} = \{1, \ldots, K\}$ with $K \in \mathbb{N}$ semantic classes. Interesting areas $\xi_I$ are given by a user-defined set of interesting features $\mathcal{F}_I \subseteq \mathcal{F}$ with $\vert \mathcal{F}_I \vert \leq K$. As the map belief $\hat{F}_t \sim p(F \mid z_{1:t}, \mathbf{p}_{1:t})$ is represented using occupancy grid maps, the unified belief $p\big(F(\mathbf{x})~\in~\mathcal{F}_{I} \mid \hat{F}_t\big)$ over interesting areas is defined as
\begin{equation} \label{eq:roi_belief_discrete}
    p\big(F(\mathbf{x}) \in \mathcal{F}_{I} \mid \hat{F}_t\big) = \sum_{f_I \in \mathcal{F_I}} p\big(F(\mathbf{x}) = f_I \mid z_{1:t}, \mathbf{p}_{1:t}\big)\,,
\end{equation}
where $f_I$ is a single class in the set of interesting classes $\mathcal{F}_I$ and $p\big(F(\mathbf{x}) = f_I \mid z_{1:t}, \mathbf{p}_{1:t}\big)$ is given by the $f_I$-th layer of the occupancy map at the grid cell corresponding to $\mathbf{x} \in \xi$, defining the categorical distribution over all $K$ classes.

Next, consider \textit{continuous feature spaces} \mbox{$\mathcal{F} = [f_a, f_b]$} with \mbox{$f_a \leq f_b$}. Interesting areas are given by user-defined thresholds \mbox{$f_{th}$ with $f_a \leq f_{th} \leq f_b$}, such that \mbox{$\mathcal{F}_I = [f_{th}, f_b]$}. As the map belief $\hat{F}_t$ is represented by Gaussian processes or Kalman filters, the probability density over feature values is given by $F(\mathbf{x}) \sim \mathcal{N}(\mu(\mathbf{x}), \sigma(\mathbf{x})^2 \mid \hat{F}_t)$ with mean~$\mu(\mathbf{x})$ and variance $\sigma(\mathbf{x})^2$ of $\hat{F}_t$ at point $\mathbf{x}$. The unified belief $p\big(F(\mathbf{x}) \in \mathcal{F}_{I} \mid \hat{F}_t\big)$ over interesting areas is defined as
\begin{equation} \label{eq:roi_belief_continuous}
    \begin{aligned}
        &p\big(F(\mathbf{x}) \in \mathcal{F}_{I} \mid \hat{F}_t\big) \\
        &= \frac{1}{\sqrt{2\pi\sigma(\mathbf{x})^2}} \int_{f_{th}} \exp\left(-\frac{(f - \mu(\mathbf{x}))^2}{2\sigma(\mathbf{x})^2}\right) \,df \\
        &= 1 - \Phi \left(\frac{f_{th} - \mu(\mathbf{x})}{\sqrt{\sigma(\mathbf{x})^2}}\right),
    \end{aligned}
\end{equation}
where $\Phi(\cdot)$ is the cumulative distribution function of the standard normal distribution measuring $p\big(F(\mathbf{x}) \leq f_{th} \mid \hat{F}_t\big)$.  

The mission-specific hyperparameters $\mathcal{H} = \{\mathcal{F}, \mathcal{F}_I, \theta_F\}$ directly influence the computation of our unified belief over interesting areas $\hat{F}_{I, t}$ in \Cref{eq:roi_belief_discrete} or \Cref{eq:roi_belief_continuous}, making the effect of the chosen mission hyperparameters accessible to the planning policy, thus improving adaptivity to the concrete instance of \Cref{eq:ipp_problem_as_rl} a planning method aims to solve. For learning-based planning methods aiming to train a policy~$\pi^*$ offline, we additionally condition the planning policy on the mission-specific hyperparameters as it allows us to train a single policy that can solve \Cref{eq:ipp_problem_as_rl} for various terrain monitoring variants $\mathcal{H}$ without retraining. 

\subsection{Adaptive IPP Reward Function}
\label{subsec:reward_function}

We introduce a new reward function for the general adaptive \ac{IPP} terrain monitoring problem in \Cref{eq:ipp_problem_as_rl} based on our unified planning state space formulation $s_t$ presented in \Cref{subsec:unified_state_space}. The unified planning state space and reward function could be integrated into any non-learning-based planning method searching for the optimal policy $\pi^*$ online or learning-based planning method for training $\pi^*$ offline. 

In adaptive \ac{IPP} problems, we aim to quickly find initially unknown areas of interest $\xi_I$ (\Cref{eq:roi}) and precisely estimate the terrain feature field $F$ in these areas. To this end, we aim to maximise information about the map belief \mbox{$\hat{F}_t \sim p(F \mid z_{1:t}, \mathbf{p}_{1:t})$} in unknown areas of interest $\xi_I$ (\Cref{eq:map_belief}). To adapt paths online towards areas likely of interest, we reward uncertainty reduction of map belief $\hat{F}_t$ proportionally to our unified belief over interesting areas $\hat{F}_{I, t}(\mathbf{x}) \sim p\big(F(\mathbf{x}) \in \mathcal{F}_{I} \mid \hat{F}_t\big)$ (\Cref{eq:roi_belief_discrete}, \Cref{eq:roi_belief_continuous}). Assume $\mathcal{X}$ is a finite subset of points $\mathbf{x} \in \xi$ sampled from an equidistant grid over the terrain $\xi$. The reward in \Cref{eq:ipp_problem_as_rl} is defined as
\begin{equation} \label{eq:reward_function}
    \begin{split}
        &R(s_t, \mathbf{a}_t, s_{t+1}) = \\
        &\sum_{\mathbf{x} \in \mathcal{X}} \frac{H(\hat{F}_{t}(\mathbf{x})) - H(\hat{F}_{t+1}(\mathbf{x}))}{H(\hat{F}_{t}(\mathbf{x}))} p\big(F(\mathbf{x}) \in \mathcal{F}_{I} \mid \hat{F}_t\big),
    \end{split}
\end{equation}
where $H(\hat{F}_{t}(\mathbf{x}))$ is the uncertainty of the mission-specific map belief $\hat{F}_t$ at a point $\mathbf{x} \in \mathcal{X}$ and $\hat{F}_{t+1}$ is the updated map belief after executing action $\mathbf{a}_t$ and collecting a new observation \mbox{$z_{t+1}~\sim~p(z \mid \mathbf{p}_{t+1}, F)$} from a next pose $\mathbf{p}_{t+1}$. For occupancy maps, $H\big(\hat{F}_t(\mathbf{x}_t)\big)$ is the exponential Shannon entropy at $\mathbf{x}$. For Gaussian processes and Kalman filters, $H\big(\hat{F}_t(\mathbf{x}_t)\big)$ is the variance trace at $\mathbf{x}$.

Assume two points $\mathbf{x}, \mathbf{x}^\prime \in \xi$. If both points have the same probability of belonging to areas of interest~$\xi_I$, the reward favours points~$\mathbf{x}$ with higher expected map uncertainty reduction to foster exploration. If both points' expected map uncertainty reduction is the same, the reward favours point $\mathbf{x}$ with a higher probability of belonging to areas of interest $\xi_I$ to focus on these areas as they are discovered. By definition of \Cref{eq:roi}, our reward also contains pure terrain exploration scenarios with areas of interest $\xi_I=\xi$ covering the whole terrain as a special case if all feature values $\mathcal{F} = \mathcal{F}_I$ are of interest. In these cases, $p(F(\mathbf{x}) \in \mathcal{F}_{I} \mid \hat{F}_t) = 1$ for all $\mathbf{x}, \mathbf{x}^\prime \in \xi$ by definition of \Cref{eq:roi_belief_discrete} and \Cref{eq:roi_belief_continuous}. Thus, points $\mathbf{x}$ with higher expected map uncertainty reduction are favoured, fostering the exploration of the whole terrain.

%Alternatively, while rewarding the uncertainty reduction of our new unified belief over areas of interest $\hat{F}_{I, t}$ would be sufficient to achieve competitive adaptive \ac{IPP} performance in many terrain monitoring missions, it is necessary to reward uncertainty reduction of the mission-specific map belief $\hat{F}_t$ in areas likely of interest. As an example, consider pure terrain exploration missions with interesting features $\mathcal{F}_I = \mathcal{F}$, such that interesting areas $\xi_I = \xi$ cover the whole terrain by definition of \Cref{eq:roi}. Assuming discrete-valued occupancy maps, $p(F(\mathbf{x}) \in \mathcal{F}_{I} \mid \hat{F}_t) = 1$ by definition of \Cref{eq:roi_belief_discrete}. Thus, $H(\hat{F}_{I,t}(\mathbf{x})) = 0$ even before a mission starts, hence not providing informative rewards. Instead, we reward uncertainty reduction over the map belief $\hat{F}_t$ proportionally to our new unified belief over interesting areas $\hat{F}_{I, t}$. This way, our reward function encourages exploring and estimating interesting areas while adaptively refining the map belief in these areas as they are discovered during a mission.

\subsection{Planning Policy Training Details} \label{subsec:policy_training}

We use \ac{RL} to train a single unified planning policy $\pi^*$ on simulated terrain monitoring deployments with previously unmet mission variations. We detail our terrain monitoring mission simulations, encoding of mission hyperparameters~$\mathcal{H}$ and the used \ac{RL} algorithm and policy network representing~$\pi^*$. In practice, any policy learning method, e.g. imitation learning, policy network architecture, and hyperparameter encoding could be used to train the policy with our new adaptive \ac{IPP} formulation introduced in \Cref{subsec:unified_state_space} and \Cref{subsec:reward_function}.

\begin{figure*}[!t]
    \begin{minipage}[t]{.73\linewidth}
        \vspace{0mm}
        \centering
        \captionof{table}{Comparison of state-of-the-art map-specifically designed and trained methods to our map-agnostic planning policy (\textit{RL-Ours}) on simulated continuous- and discrete-valued terrain feature monitoring missions. Best average performances are marked in bold, second-best average performances are underlined if standard deviations in brackets overlap. Our map-agnostic policy performs best in case of \textit{Varying} user-defined mission hyperparameters and similar to state-of-the-art adaptive \ac{IPP} methods in case of \textit{Static} mission hyperparameters.}
        \label{T:sim_results}
        \small
        \resizebox{\linewidth}{!}{%
        \tabcolsep=0.1cm
        \begin{tabular}{
            @{}>{\centering\arraybackslash}m{0.2cm}
            >{\centering\arraybackslash}m{1.6cm}
            >{\centering\arraybackslash}m{5.5cm}
            >{\centering\arraybackslash}m{5.5cm}
            >{\centering\arraybackslash}m{1.6cm}@{}}
        \toprule
        & \textbf{Approach} & \textbf{Static $\mathcal{H}$} & \textbf{Varying $\mathcal{H}$} & \textbf{Replanning} \\ 
        & & \enskip\enskip\enskip II$\uparrow$ \enskip\enskip\enskip Unc.$\downarrow$ \enskip\enskip MLL$\downarrow$ \enskip\enskip RMSE$\downarrow$ & \enskip\enskip\enskip II$\uparrow$ \enskip\enskip\enskip Unc.$\downarrow$ \enskip\enskip MLL$\downarrow$ \enskip\enskip RMSE$\downarrow$ & \enskip \textbf{time} [s]$\downarrow$ \\
        \midrule
        \multirow{7}{*}{\rotatebox[origin=c]{90}{Continuous}}
            & RL-Ours & \underline{25.8} {\scriptsize(0.17)} \underline{60.6} {\scriptsize(0.22)} -64.6 {\scriptsize(0.12)} \underline{3.83} {\scriptsize(0.08)} & \textbf{26.2} {\scriptsize(0.65)} \textbf{60.4} {\scriptsize(0.64)} \textbf{-60.5} {\scriptsize(0.59)} \underline{3.81} {\scriptsize(0.07)} & \textbf{0.004} \\
            & RL-Base-C & \textbf{26.1} {\scriptsize(0.25)} \textbf{59.6} {\scriptsize(0.28)} \textbf{-66.2} {\scriptsize(0.39)} \textbf{3.67} {\scriptsize(0.02)} & 24.0 {\scriptsize(0.94)} 64.2 {\scriptsize(1.07)} -48.4 {\scriptsize(4.64)} 5.51 {\scriptsize(0.97)} & \textbf{0.004} \\ \cmidrule(l){2-5}
            & MCTS & 25.6 {\scriptsize(0.09)} 60.7 {\scriptsize(0.08)} \underline{-64.9} {\scriptsize(0.52)} \underline{3.83} {\scriptsize(0.16)} & \underline{25.3} {\scriptsize(0.41)} \underline{61.1} {\scriptsize(0.24)} \underline{-59.5} {\scriptsize(0.70)} 4.27 {\scriptsize(0.22)} & 2.86 \\
            & CMA-ES & 23.1 {\scriptsize(1.27)} 63.0 {\scriptsize(3.41)} -60.1 {\scriptsize(6.38)} 5.45 {\scriptsize(2.67)} & 21.5 {\scriptsize(1.44)} 64.3 {\scriptsize(2.32)} -55.4 {\scriptsize(5.52)} \textbf{2.59} {\scriptsize(0.85)} & 6.05 \\
            & Greedy & 24.5 {\scriptsize(0.14)} 62.0 {\scriptsize(0.12)} -62.0 {\scriptsize(0.26)} 4.11 {\scriptsize(0.11)} & 25.2 {\scriptsize(0.66)} 61.7 {\scriptsize(0.29)} -58.0 {\scriptsize(1.55)} 4.35 {\scriptsize(0.24)} & 0.05 \\
            & Coverage & 15.3 {\scriptsize(0.16)} 75.5 {\scriptsize(0.39)} -28.0 {\scriptsize(1.44)} 10.8 {\scriptsize(0.12)} & 13.4 {\scriptsize(0.28)} 77.1 {\scriptsize(0.25)} -30.6 {\scriptsize(0.87)} 9.93 {\scriptsize(0.20)} & - \\ \midrule \midrule
            & & \enskip II$\uparrow$ \enskip\enskip\enskip Unc.$\downarrow$ \enskip\enskip mIoU$\uparrow$ \enskip\enskip\enskip F1$\uparrow$ & \enskip II$\uparrow$ \enskip\enskip\enskip Unc.$\downarrow$ \enskip\enskip mIoU$\uparrow$ \enskip\enskip\enskip F1$\uparrow$ & \\ \cmidrule(l){3-5}
        \multirow{7}{*}{\rotatebox[origin=c]{90}{Discrete}} 
            & RL-Ours & \textbf{31.5} {\scriptsize(0.12)} \textbf{38.3} {\scriptsize(0.76)} \textbf{20.8} {\scriptsize(0.26)} \textbf{25.6} {\scriptsize(0.19)} & \underline{30.5} {\scriptsize(0.12)} \textbf{39.2} {\scriptsize(0.42)} \textbf{20.4} {\scriptsize(0.17)} \textbf{25.3} {\scriptsize(0.12)} & \textbf{0.004} \\
            & RL-Base-D & \underline{31.1} {\scriptsize(0.09)} \underline{38.6} {\scriptsize(0.43)} \underline{20.7} {\scriptsize(0.14)} \underline{25.5} {\scriptsize(0.09)} & 30.4 {\scriptsize(0.57)} \underline{40.2} {\scriptsize(0.95)} \underline{20.1} {\scriptsize(0.29)} \underline{25.0} {\scriptsize(0.26)} & \textbf{0.004} \\ \cmidrule(l){2-5}
            & MCTS & 30.7 {\scriptsize(0.25)} 41.9 {\scriptsize(0.31)} 19.5 {\scriptsize(0.08)} 24.5 {\scriptsize(0.12)} & \textbf{30.8} {\scriptsize(0.21)} 41.2 {\scriptsize(0.85)} 19.8 {\scriptsize(0.31)} 24.7 {\scriptsize(0.24)} & 1.95 \\
            & CMA-ES & 29.6 {\scriptsize(1.87)} 43.2 {\scriptsize(2.38)} 19.2 {\scriptsize(0.85)} 24.2 {\scriptsize(0.76)} & 30.0 {\scriptsize(1.45)} 42.4 {\scriptsize(0.61)} 19.5 {\scriptsize(0.42)} 24.4 {\scriptsize(0.51)} & 3.75 \\
            & Greedy & 29.9 {\scriptsize(0.24)} 44.4 {\scriptsize(0.83)} 18.7 {\scriptsize(0.26)} 23.8 {\scriptsize(0.22)} & 29.4 {\scriptsize(0.17)} 45.6 {\scriptsize(0.59)} 18.2 {\scriptsize(0.22)} 23.2 {\scriptsize(0.22)} & 0.03 \\
            & Coverage & 29.7 {\scriptsize(0.46)} 44.3 {\scriptsize(0.37)} 18.7 {\scriptsize(0.12)} 23.8 {\scriptsize(0.12)} & 27.9 {\scriptsize(0.24)} 45.3 {\scriptsize(0.37)} 18.4 {\scriptsize(0.12)} 23.3 {\scriptsize(0.08)} & - \\ \bottomrule
        \end{tabular}%
        }
    \end{minipage}
    \hfill
    \begin{minipage}[t]{.24\linewidth}
        \vspace{0mm}
        \centering
        \captionsetup[subfloat]{labelformat=empty, skip=0.5mm}
        \subfloat[Ground truth]{\includegraphics[width=0.49\columnwidth]{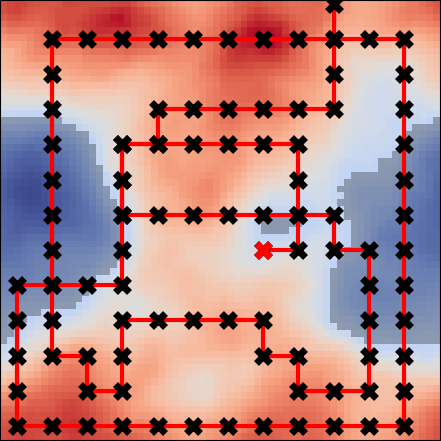}}\hfill
        \subfloat[Ground truth]{\includegraphics[width=0.49\columnwidth]{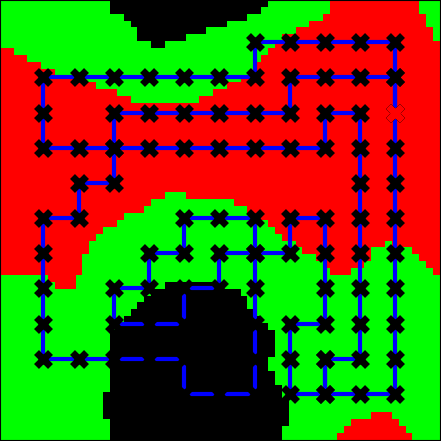}}
        \vspace{0mm}
        \subfloat[Map belief]{\includegraphics[width=0.49\columnwidth]{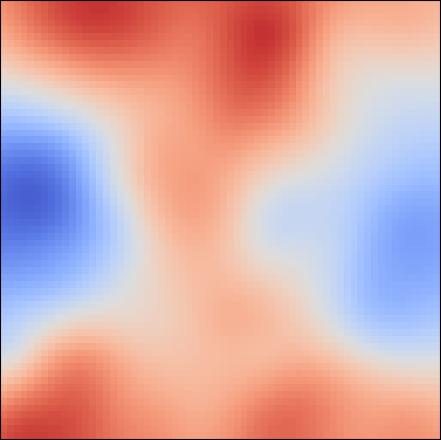}}\hfill
        \subfloat[Map belief]{\includegraphics[width=0.49\columnwidth]{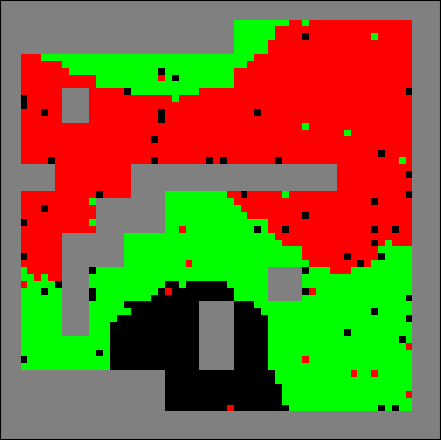}}
        \vspace{0.25mm}
        \subfloat[Unified belief]{\includegraphics[width=0.49\columnwidth]{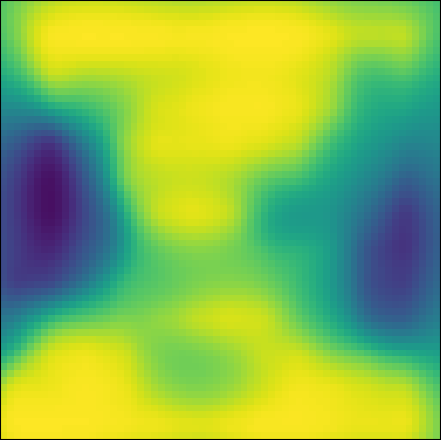}}\hfill
        \subfloat[Unified belief]{\includegraphics[width=0.49\columnwidth]{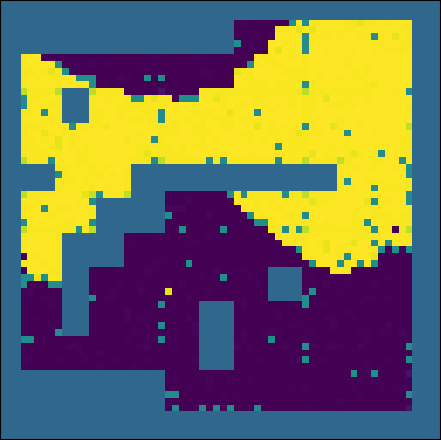}}
        \vspace{-1mm}
        \captionof{figure}{Continuous (left) and discrete terrain feature fields (right).}
        \label{F:exp_qualititative_results}
    \end{minipage}
    \vspace{-3mm}
\end{figure*}

\textbf{Mission simulations.} We sample mission hyperparameters $\mathcal{H} = \{\mathcal{F}, \mathcal{F}_I, \theta_F\}$, defining the monitoring mission. We randomly choose continuous- or discrete-valued terrain features~$\mathcal{F}$. Note that we train our policy on both classes of terrain feature monitoring missions to minimise the training to deployment gap~\cite{kirk2023survey} and maximise planning performance. In case of continuous-valued features, we use a Gaussian process with sampled kernel parameters $\theta_F$ to represent the map belief $\hat{F}_t$. In case of discrete-valued features, we use an occupancy map to represent $\hat{F}_t$. For given features, we simulate randomised ground truth feature fields $F$ with spatial correlations of different extents as depicted in \Cref{F:exp_qualititative_results}. 
%We sample the set of interesting feature values $\mathcal{F}_I$ from feature values $\mathcal{F}$ and the set of map parameters $\theta_F = \{l_{GP}\}$ uniformly at random.

\textbf{Hyperparameter encoding.} We explicitly input mission hyperparameters $l_{GP}$ and $f_{th}$ into state $s_t$. The map hyperparameter $l_{GP} \geq 0 \in \theta_F$ is the lengthscale of a Gaussian process Matern kernel used to represent the map belief~$\hat{F}_t$. This is important as different lengthscales result in different map updates along paths, potentially affecting decision-making. Map beliefs $\hat{F}_t$ assuming spatially independent measurements $z$, e.g. occupancy grid maps, are naturally encoded by $l_{GP} = 0$ as Matern kernels with \mbox{$l_{GP} \to 0$} assume spatially independent measurements. The user-defined value threshold $f_{th} \in \mathcal{F}$ represents the interesting features $\mathcal{F}_I$. 
%Interesting feature thresholds $f_{th}$ explicitly encode prior user belief about the spatial extent of interesting areas $\xi_I$ over the terrain $\xi$. For example, for continuous- or discrete-valued exploration missions with $\mathcal{F}_I = \mathcal{F}$, conditioning the policy on $f_{th} = 0$ encodes those interesting areas cover the whole terrain.

\textbf{Policy training.} We train our policy $\pi^*$ using the \acl{PPO} algorithm~\cite{schulman2017proximal} and compute our state space in \Cref{eq:env_state} over an equidistant grid $\mathcal{X}_t$. We use the IMPALA encoder~\cite{espeholt2018impala} to process the interesting area belief $p\big(F(\mathbf{x})~\in~\mathcal{F}_{I} \mid \hat{F}_t\big)$ and map belief uncertainty $H(\hat{F}_{t}(\mathbf{x}))$ for each $\mathbf{x} \in \mathcal{X}_t$. We use a \ac{MLP} to process the current robot's pose $\mathbf{p}_t$, remaining budget $B_t$ and mission hyperparameters $\mathcal{H}$, and a \ac{MLP} head to predict the stochastic policy $\pi(s_t)$ and value function $V_\pi(s_t)$.

% \textbf{Reward function scaling.} As hyperparameters $\mathcal{H}$ vary in each mission, the reward magnitudes vary between missions, leading to unstable policy training. To this end, we rescale the reward in \Cref{eq:reward_function} during training. We normalise map belief uncertainties $H(\hat{F}_t)$, $H(\hat{F}_{t+1})$ by the prior map belief uncertainty $H(\hat{F}_0)$. Further, as the interesting feature threshold $f_{th}$ increases, the map belief uncertainty reduction in interesting areas decreases as they cover less terrain area. Similarly, as the Gaussian process kernel length scale $l_{GP}$ decreases, the map belief uncertainty reduction decreases as the assumed spatial dependence of measurements decreases. Thus, we exponentially scale the received reward with decreasing kernel length scale $l_{GP}$ and linearly scale the reward with increasing interesting feature threshold $f_{th}$.
	
%===============================================================================

\section{Experimental Results}
\label{sec:results}

%Our experiments are designed to answer two main research questions. First, can we train a single adaptive \ac{IPP} policy for monitoring missions with varying terrain map representations and user-defined mission characteristics using our proposed map-agnostic adaptive \ac{IPP} formulation? Second, can we integrate our map-agnostic adaptive \ac{IPP} formulation into state-of-the-art online non-learning-based policy search methods without performance loss? To answer these questions, in \Cref{subsec:exp_simulation_results}, we show that training our single unified policy on various monitoring missions yields competitive performance with state-of-the-art online non-learning-based policy search methods and offline-learned policies adapted and re-trained for each class of monitoring missions with specific terrain map representation. In \Cref{subsec:exp_real_world_dataset_results}, we verify that our single unified policy trained in simulation performs similarly to these state-of-the-art adaptive \ac{IPP} methods on unseen real-world datasets. In \Cref{subsec:ablation_study}, we show that our map-agnostic adaptive \ac{IPP} formulation unifies existing adaptive \ac{IPP} methods while maintaining or improving planning performance.
The experiments are designed to support our claims. In \Cref{subsec:exp_simulation_results}, we show that training our map-agnostic policy on various monitoring missions yields competitive performance with state-of-the-art online non-learning-based policy search methods and offline-learned policies adapted and re-trained for each class of monitoring missions with specific terrain map representation. In \Cref{subsec:exp_real_world_dataset_results}, we verify that our map-agnostic policy trained in simulation performs similarly to these state-of-the-art adaptive \ac{IPP} methods on unseen real-world datasets. In \Cref{subsec:ablation_study}, we show that our map-agnostic adaptive \ac{IPP} formulation unifies existing adaptive \ac{IPP} methods while maintaining or improving their performance. 
%Further, we perform an ablation study on different reward functions and show that our new reward function yields the best performance for training a single unified policy.

\subsection{Experimental Setup}
\label{subsec:exp_setup}

\textbf{Mission setup.} The general procedure for simulating monitoring missions used to train and evaluate planning policies is described in \Cref{subsec:policy_training}. For discrete-valued terrain features, we assume three semantic classes $\mathcal{F}$ with interesting classes $\mathcal{F}_I$ of varying spatial extent. We equip a simulated \ac{UAV} with a sensor delivering image-like semantic measurements $z_t$ spanning a downwards-projected field of view. We use occupancy grid maps $\hat{F}_t$ for terrain mapping and confusion matrix-based sensor noise as in~\cite{popovic2020informative, westheider2023multi}. For continuous-valued terrain features, we assume features $\mathcal{F} = [0,1]$ with interesting thresholds $f_{th}$, such that $\mathcal{F}_I = [f_{th}, 1]$. Simulated \acp{UAV} are equipped with sensors delivering point measurements $z_t$ with Gaussian noise, mapped using Gaussian processes as in ~\cite{hitz2017adaptive, popovic2020informative, cao2023catnipp, ott2023sboaippms}. We distinguish between the classical evaluation protocol of fixed mission hyperparameters \mbox{$\mathcal{H} = \{f_{th}, l_{GP}\} = \{0.4, 0.35\}$} as in~\cite{hitz2017adaptive, popovic2020informative,ruckin2022adaptive, cao2023catnipp}, denoted as \textit{Static}, and our more challenging scenario of randomly sampled $\mathcal{H}$ with $f_{th} \in [0.0, 0.8]$ and $l_{GP} \in [0.15, 0.55]$ denoted below as \textit{Varying}. This resembles the static mission hyperparameters in expectation. The initial mission budget is set to $B = 100$s, and initial robot positions~$\mathbf{p}_0$ are sampled at random. We assume actions \mbox{$\mathbf{a}_t \in \mathcal{A}$} representing relative 2D robot position changes on an equidistant grid as in~\cite{ott2023sboaippms, popovic2020informative, ruckin2022adaptive, westheider2023multi}. To train and benchmark our approach, we simulate ground truth feature fields $F$ with varying spatial correlations as shown in \Cref{F:exp_qualititative_results}-top. Additionally, we evaluate the performance on real-world orthomosaic fields $F$.

\textbf{Baselines.} We consider state-of-the-art adaptive \ac{IPP} methods performing online planning or offline-trained policy inference. In contrast to our \textit{RL-Ours} method, all baseline policies rely on map-specific planning state spaces. All methods consider the current robot position and remaining budget in their state. Continuous-valued terrain features are modelled by directly using posterior mean and variance of the Gaussian process as in~\cite{hitz2017adaptive, popovic2020informative, ott2023sboaippms, ruckin2022adaptive, cao2023catnipp}. Discrete-valued terrain features are modelled by directly using the posterior occupancy map and its entropy as in~\cite{popovic2020informative, westheider2023multi}. All baselines reward map uncertainty reduction in approximated areas of interest relying on hand-tuned confidence intervals as in~\cite{popovic2020informative, cao2023catnipp, hitz2017adaptive}. Based on these states and rewards, we implement finite-horizon rollout-based \textit{\ac{MCTS}}~\cite{choudhury2020adaptive, ott2023sboaippms}, finite-length path optimisation using the \textit{\ac{CMA-ES}}~\cite{hitz2017adaptive, popovic2020informative}, and \textit{Greedy} planning~\cite{popovic2020informative} as online policy search methods. To offline-train \textit{RL-Base} planning policies, we use \ac{RL} assuming \textit{Static} hyperparameters and perform policy inference online~\cite{westheider2023multi, cao2023catnipp, ruckin2022adaptive, niroui2019deep}. Further, we pre-compute lawnmower-like \textit{Coverage} paths~\cite{galceran2013survey} commonly used in real-world monitoring deployments.

\textbf{Evaluation metrics.} All adaptive replanning performance metrics are computed over areas of interest $\xi_I$ (\Cref{eq:roi}) after a mission is terminated. For continuous-valued mapping missions, we compute the final covariance $\log$-trace of map~$\hat{F}_t$ normalised by the prior covariance $\log$-trace of $\hat{F}_0$ (Unc.) and \ac{RMSE} as in \cite{popovic2020informative, cao2023catnipp, ott2023sboaippms}, and \ac{MLL} of $\hat{F}_t$ w.r.t. the ground truth feature field $F$ as computed by Marchant and Ramos~\cite{marchant2014icra}, Eq. (23). For discrete-valued mapping missions, we compute the final Shannon entropy of map $\hat{F}_t$ normalised by the prior Shannon entropy of $\hat{F}_0$ (Unc.), and \ac{mIoU} and F1-score of $\hat{F}_t$ w.r.t. the ground truth feature field $F$ as in~\cite{popovic2020informative, westheider2023multi, niroui2019deep}. Further, we compute an \ac{II} as one minus the area under the normalised map uncertainty (Unc.) over budget curve. The \ac{II} captures the uncertainty reduction speed over the depleted budget in a single metric. All metrics are averaged over $100$ missions, repeated with three different random seeds. We report mean and standard deviations over the three seeds.

\subsection{Simulation Results}
\label{subsec:exp_simulation_results}

The first set of experiments shows that our single map-agnostic adaptive \ac{IPP} policy yields competitive performance with state-of-the-art online policy search methods while substantially reducing replanning runtime. Further, our map-agnostic policy outperforms state-of-the-art map-specifically designed and offline-trained policies on various terrain monitoring missions. We evaluate all methods in simulated continuous- and discrete-valued terrain feature monitoring scenarios as described in \Cref{subsec:exp_setup}. We consider the classical \textit{Static} mission hyperparameter and our \textit{Varying} mission hyperparameter evaluation protocol to benchmark adaptive \ac{IPP} approaches on challenging inter-mission variations.

\Cref{T:sim_results} summarises the results. In line with previous \ac{RL}-based adaptive \ac{IPP} works, map-specifically designed \textit{RL-Base-C} and \textit{RL-Base-D} policies outperform state-of-the-art online policy search methods in their respective continuous- and discrete-valued terrain feature monitoring missions with \textit{Static} hyperparameters they were trained on. Our single map-agnostic \textit{RL-Ours} policy shows competitive performance on continuous- and discrete-valued monitoring missions with \textit{Static} hyperparameters compared to online policy search methods and the \textit{RL-Base-C/D} policies. Noticeably, our map-agnostic policy outperforms the map-specific \textit{RL-Base-C/D} policies on \textit{Varying} hyperparameters, causing larger inter-mission variations. This verifies the advantage of our unified policy being trained and conditioned on larger mission variations, while the \textit{RL-Base-C} trained on \textit{Static} hyperparameters does not match the performance of online policy search. Further, our map-agnostic policy outperforms the strongest \textit{\ac{MCTS}} adaptive \ac{IPP} method on missions with \textit{Varying} hyperparameters while substantially reducing replanning runtimes at deployment. This shows we can successfully train a single adaptive \ac{IPP} policy applicable and well-performing in monitoring scenarios with larger inter-mission variations in user-defined hyperparameters and terrain map representations. \Cref{F:exp_qualititative_results} shows simulated ground truth feature fields $F$ with paths planned based on our unified belief $\hat{F}_{I,t}$ over initially unknown non-shaded and red areas of interest $\xi_I$, derived from mission-specific map beliefs $\hat{F}_t$ with yellow indicating a high probability of interesting areas according to $\hat{F}_{I,t}$.

\subsection{Results on Real-World Datasets}
\label{subsec:exp_real_world_dataset_results}

\begin{figure*}[!t]
    \begin{minipage}[t]{.75\linewidth}
        \vspace{0mm}
        \centering
        \captionof{table}{Comparison of state-of-the-art map-specifically designed and trained methods to our map-agnostic planning policy (\textit{RL-Ours}) on real-world continuous-valued surface temperature (\textit{Temperature-1/2}) and discrete-valued urban (\textit{Potsdam}) and rural (\textit{RIT-18}) semantic terrain datasets. Best average performances are marked in bold, second-best average performances are underlined if standard deviations in brackets overlap. Our map-agnostic policy performs similarly to state-of-the-art adaptive \ac{IPP} methods.}
        \label{T:real_world_results}
        \small
        \tabcolsep=0.0cm
        \begin{tabular}{
            @{}>{\centering\arraybackslash}m{1.8cm}
            >{\centering\arraybackslash}m{2.8cm}
            >{\centering\arraybackslash}m{2.8cm}
            >{\centering\arraybackslash}m{2.8cm}
            >{\centering\arraybackslash}m{2.8cm}@{}}
        \toprule
        \textbf{Approach} & \textbf{Temperature-1} & \textbf{Temperature-2} & \textbf{Potsdam}~\cite{Potsdam2018} & \textbf{RIT-18}~\cite{kemker2018algorithms} \\ 
        & II$\uparrow$ \enskip\enskip\enskip Unc.$\downarrow$ & II$\uparrow$ \enskip\enskip\enskip Unc.$\downarrow$ & II$\uparrow$ \enskip\enskip\enskip Unc.$\downarrow$ & II$\uparrow$ \enskip\enskip\enskip Unc.$\downarrow$ \\ \midrule
        RL-Ours & \textbf{25.4} {\scriptsize(0.31)} \underline{62.2} {\scriptsize(0.09)} & \underline{27.6} {\scriptsize(0.26)} \underline{58.6} {\scriptsize(0.22)} & \textbf{32.9} {\scriptsize(1.92)} \textbf{35.8} {\scriptsize(3.10)} & \underline{31.7} {\scriptsize(1.25)} \underline{39.9} {\scriptsize(0.64)} \\ 
        RL-Base & 24.6 {\scriptsize(0.62)} 63.1 {\scriptsize(0.46)} & 26.7 {\scriptsize(0.14)} 60.6 {\scriptsize(0.85)} & \underline{31.9} {\scriptsize(1.82)} \underline{36.4} {\scriptsize(2.24)} & \textbf{32.2} {\scriptsize(0.31)} \textbf{39.6} {\scriptsize(0.59)} \\ \midrule
        \ac{MCTS} & \underline{25.3} {\scriptsize(0.22)} \textbf{61.9} {\scriptsize(0.21)} & \textbf{27.7} {\scriptsize(0.49)} \textbf{58.4} {\scriptsize(0.49)} & 31.7 {\scriptsize(0.45)} 40.8 {\scriptsize(0.93)} & 30.7 {\scriptsize(0.50)} 42.2 {\scriptsize(0.17)} \\
        Greedy & 25.2 {\scriptsize(0.46)} 62.4 {\scriptsize(0.45)} & 27.1 {\scriptsize(0.43)} 58.9 {\scriptsize(0.29)} & 29.8 {\scriptsize(0.29)} 44.0 {\scriptsize(0.66)} & 29.7 {\scriptsize(0.45)} 46.0 {\scriptsize(0.54)} \\
        Coverage & 14.7 {\scriptsize(0.34)} 75.9 {\scriptsize(0.08)} & 15.8 {\scriptsize(0.84)} 73.4 {\scriptsize(0.98)} & 29.3 {\scriptsize(0.60)} 44.6 {\scriptsize(0.17)} & 29.7 {\scriptsize(0.62)} 45.4 {\scriptsize(0.91)} \\ \bottomrule
        \end{tabular}
    \end{minipage}
    \hfill
    \begin{minipage}[t]{.23\linewidth}
        \vspace{0mm}
        \centering
        \captionsetup[subfloat]{labelformat=empty, skip=0.5mm}
        \subfloat[Temperate-1]{\includegraphics[width=0.49\columnwidth]{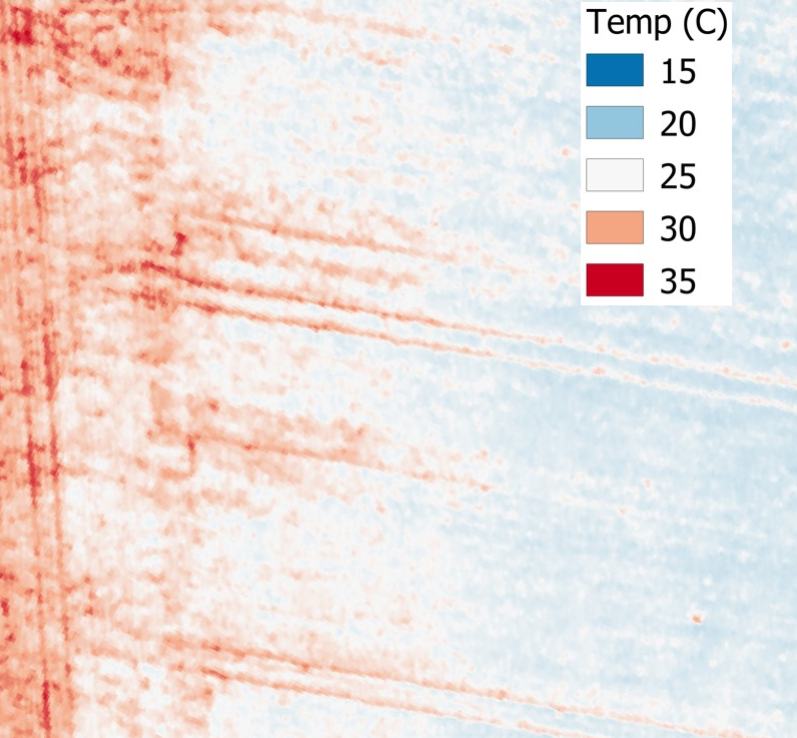}}\hfill
        \subfloat[Temperature-2]{\includegraphics[width=0.49\columnwidth]{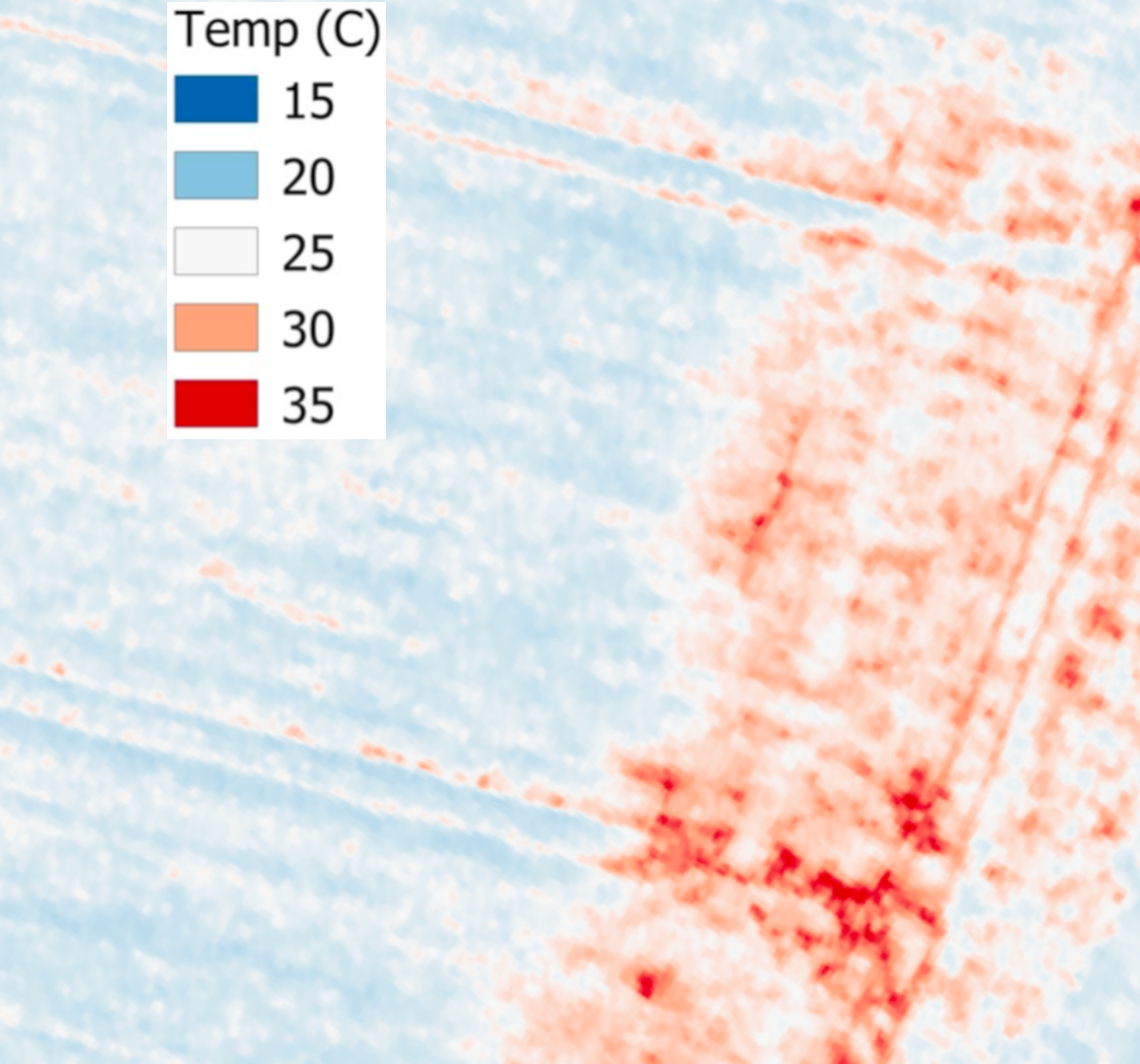}}
        \vspace{1mm}
        \subfloat[Potsdam]{\includegraphics[width=0.49\columnwidth]{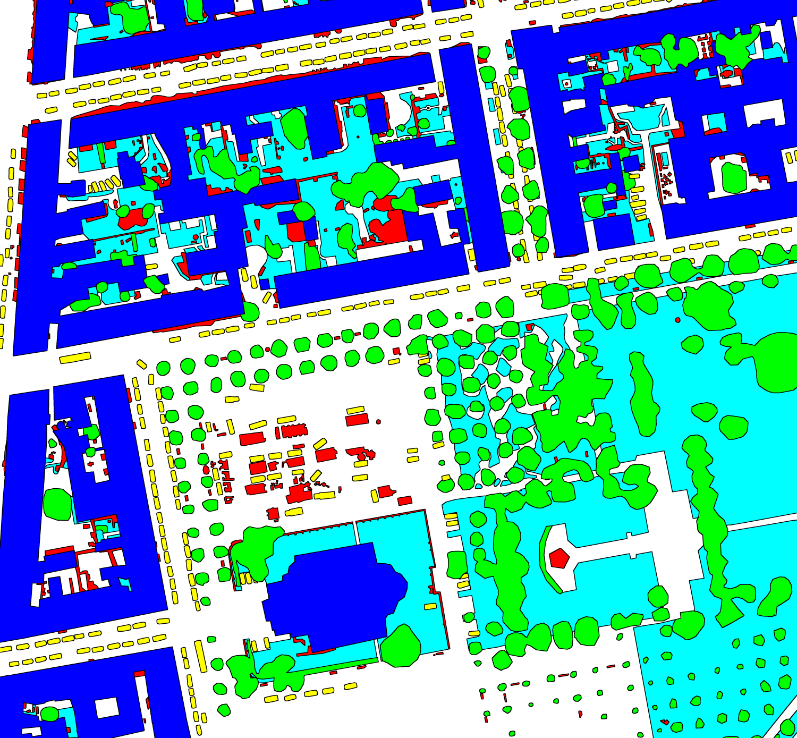}}\hfill
        \subfloat[RIT-18]{\includegraphics[width=0.49\columnwidth, height=0.46\columnwidth]{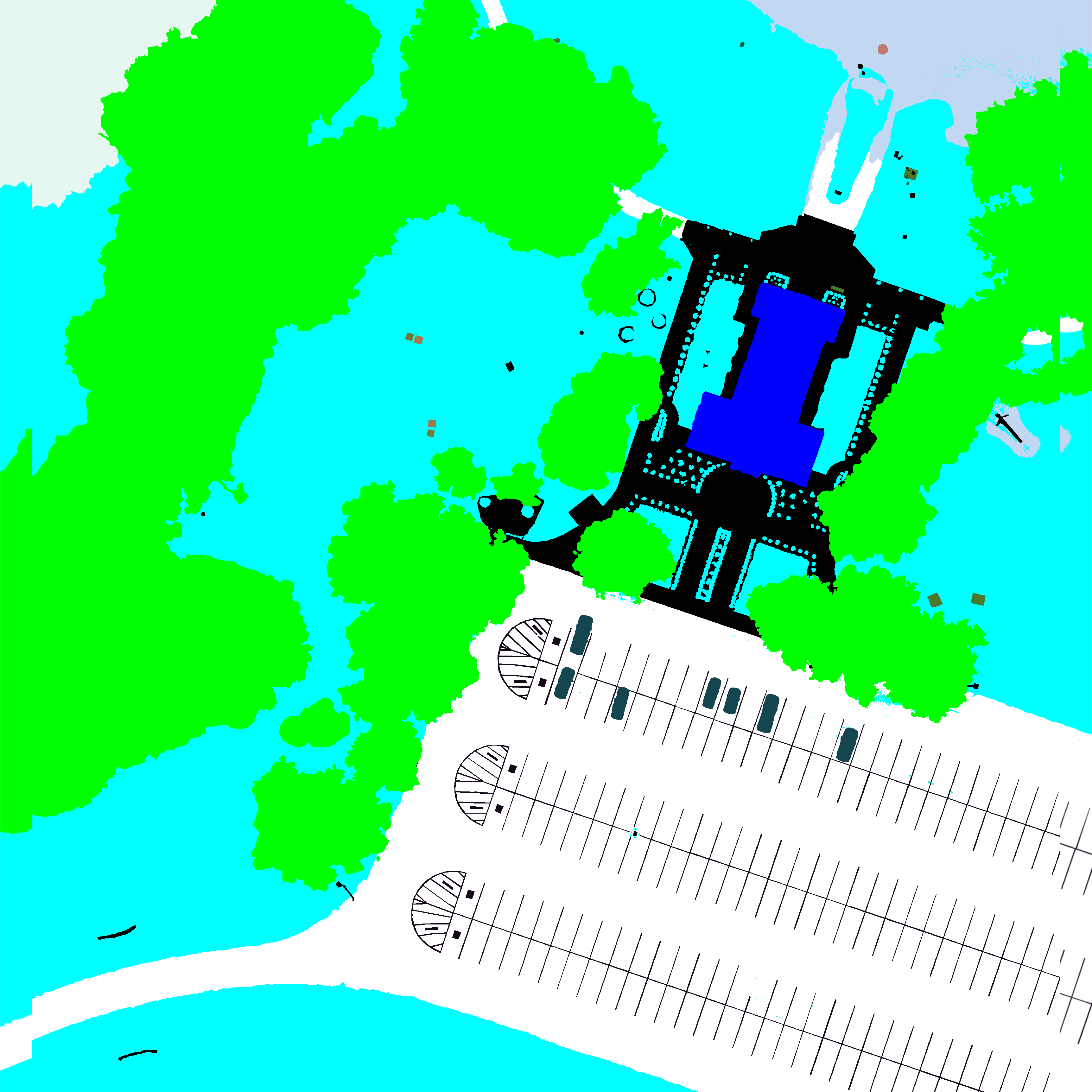}}
        \vspace{-1mm}
        \captionof{figure}{Real-world datasets.}
        %(Left) Crop fields with high surface temperatures (red). (Right) Semantic urban and rural areas with vegetation (green and light blue).}
        \label{F:exp_real_world_orthos}
    \end{minipage}
    \vspace{-2mm}
\end{figure*}

The experiments on real-world orthomosaics are designed to show that our single unified policy trained in simulation performs similarly to state-of-the-art adaptive \ac{IPP} methods on previously unseen real-world terrain datasets. We compare our map-agnostic policy (\textit{RL-Ours}) to map-specifically designed online non-learning-based planning methods and map-specifically designed and trained policies (\textit{RL-Base-C/D}). We consider two continuous-valued surface temperature orthomosaics of crop fields near Bonn, Germany, mapped using Gaussian processes, where high surface temperatures above $25^\circ$C are interesting. Further, we execute discrete-valued semantic monitoring of an urban area in Potsdam, Germany~\cite{Potsdam2018} and a rural area~\cite{kemker2018algorithms} (RIT-18), mapped using occupancy maps, where vegetation features are of interest. Orthomosaic datasets are illustrated in \Cref{F:exp_real_world_orthos}. All \ac{RL}-based policies are trained in simulation as in \Cref{subsec:exp_simulation_results} and deployed on the real-world datasets without adaptation.

\Cref{T:real_world_results} summarises the results. In line with state-of-the-art methods, our map-agnostic policy consistently outperforms traditionally used non-adaptive \textit{Coverage} paths, showcasing the advantages of adaptive online replanning. Notably, in most scenarios, our map-agnostic policy outperforms \textit{Greedy} planning and performs similarly to \textit{\ac{MCTS}} planning while substantially reducing replanning runtimes. Furthermore, our map-agnostic policy performs comparably to map-specifically designed and trained learning-based \textit{RL-Base-C/D} policies. Generally, we observe an expected small performance degradation of \ac{RL}-based policies compared to their performance in simulated missions due to simulation to real-world dataset gaps~\cite{kirk2023survey}. While our single map-agnostic policy is applied to all real-world dataset missions, each baseline requires two map-specific versions before deployment.
%Particularly, the \textit{RL-Base-D} policy is only applicable to semantic monitoring missions with the same number of classes used for training, further complicating the deployment of these learned policies. 
Overall, these results highlight the advantages of our map-agnostic policy, validating its performance on unseen real-world terrain data while facilitating deployment.

\subsection{Map-Agnostic Online Adaptive IPP Policy Search} \label{subsec:ablation_study}

The next set of experiments aims to answer if we can easily integrate our map-agnostic adaptive \ac{IPP} formulation into state-of-the-art online non-learning-based policy search methods without planning performance loss. We show that our map-agnostic adaptive \ac{IPP} formulation unifies existing adaptive \ac{IPP} methods while maintaining or improving performance in various terrain monitoring missions. 
%Further, we investigate different reward functions and verify that our new reward function yields the best performance for training a single unified policy on varying terrain monitoring missions.

\begin{table}[!t]
\centering
\caption{Integration of our map-agnostic adaptive \ac{IPP} formulation (\textit{ours}) into state-of-the-art online policy search methods. Best average performances are marked in bold, second-best average performances are underlined if standard deviations in brackets overlap. Our map-agnostic formulation unifies existing adaptive \ac{IPP} methods while consistently maintaining or improving performance over previous map-specific formulations for continuous- (\textit{prev-C}) and discrete-valued (\textit{prev-D}) terrain feature monitoring missions.}

\label{T:ablation_study_results}
\small
\tabcolsep=0.17cm
\begin{tabular}{
    @{}>{\centering\arraybackslash}m{0.1cm}
    >{\centering\arraybackslash}m{1.0cm}
    >{\centering\arraybackslash}m{0.9cm}
    >{\centering\arraybackslash}m{5.4cm}}
\toprule
& \textbf{Policy} & \textbf{IPP} & \textbf{Varying} $\mathcal{H}$ \\ 
& & & \enskip\enskip II$\uparrow$ \enskip\enskip\enskip Unc.$\downarrow$ \enskip\enskip\enskip MLL$\downarrow$ \enskip\enskip RMSE$\downarrow$ \\
\midrule
\multirow{7}{*}{\rotatebox[origin=c]{90}{Continuous}} 
    & \multirow{2}{*}{Greedy} & prev-C & \underline{25.2} {\scriptsize(0.66)} \underline{61.7} {\scriptsize(0.29)} \underline{-58.0} {\scriptsize(1.55)} \underline{4.35} {\scriptsize(0.24)} \\
    & & ours & \textbf{25.3} {\scriptsize(0.41)} \textbf{61.3} {\scriptsize(0.42)} \textbf{-59.0} {\scriptsize(0.73)} \textbf{4.15} {\scriptsize(0.16)} \\ \cmidrule(l){2-4}
    & \multirow{2}{*}{MCTS} & prev-C & 25.3 {\scriptsize(0.24)} 61.1 {\scriptsize(0.24)} -59.5 {\scriptsize(0.70)} \underline{4.27} {\scriptsize(0.22)} \\
    & & ours & \textbf{27.0} {\scriptsize(0.42)} \textbf{59.6} {\scriptsize(0.26)} \textbf{-63.8} {\scriptsize(0.34)} \textbf{4.00} {\scriptsize(0.24)} \\ \cmidrule(l){2-4} 
    & \multirow{2}{*}{CMA-ES} & prev-C & \underline{21.5} {\scriptsize(1.44)} \textbf{64.3} {\scriptsize(2.32)} \textbf{-55.4} {\scriptsize(5.52)} \textbf{2.59} {\scriptsize(0.85)} \\
    & & ours & \textbf{21.8} {\scriptsize(2.25)} \underline{64.5} {\scriptsize(1.97)} \underline{-54.1} {\scriptsize(5.67)} \underline{2.73} {\scriptsize(1.26)} \\ \midrule \midrule
    & & & \enskip\enskip II$\uparrow$ \enskip\enskip\enskip Unc.$\downarrow$ \enskip\enskip\enskip mIoU$\uparrow$ \enskip\enskip F1$\uparrow$ \\ \cmidrule(l){4-4}
\multirow{7}{*}{\rotatebox[origin=c]{90}{Discrete}} 
    & \multirow{2}{*}{Greedy} & prev-D & 29.4 {\scriptsize(0.17)} 45.6 {\scriptsize(0.59)} 18.2 {\scriptsize(0.22)} 23.2 {\scriptsize(0.22)} \\
    & & ours & \textbf{30.5} {\scriptsize(0.33)} \textbf{44.5} {\scriptsize(0.19)} \textbf{18.6} {\scriptsize(0.00)} \textbf{23.6} {\scriptsize(0.05)} \\ \cmidrule(l){2-4}
    & \multirow{2}{*}{MCTS} & prev-D & \underline{30.8} {\scriptsize(0.21)} \underline{41.2} {\scriptsize(0.85)} \textbf{19.8} {\scriptsize(0.31)} \textbf{24.7} {\scriptsize(0.24)} \\
    & & ours & \textbf{31.4} {\scriptsize(0.78)} \textbf{41.0} {\scriptsize(0.96)} \textbf{19.8} {\scriptsize(0.31)} \textbf{24.7} {\scriptsize(0.31)} \\ \cmidrule(l){2-4} 
    & \multirow{2}{*}{CMA-ES} & prev-D & \textbf{30.0} {\scriptsize(1.45)} \underline{42.4} {\scriptsize(0.61)} \underline{19.5} {\scriptsize(0.42)} \underline{24.4} {\scriptsize(0.51)} \\
    & & ours & \underline{29.6} {\scriptsize(1.33)} \textbf{41.6} {\scriptsize(0.37)} \textbf{19.7} {\scriptsize(0.21)} \textbf{24.6} {\scriptsize(0.42)} \\ \bottomrule
\end{tabular}
\vspace{-2mm}
\end{table}

To showcase the general applicability of our approach, we integrate our map-agnostic adaptive \ac{IPP} formulation (\textit{ours}) with the greedy, \ac{MCTS} and \ac{CMA-ES} algorithms described in \Cref{subsec:exp_setup} using our state formulation in \Cref{eq:env_state} and reward function in \Cref{eq:reward_function} for policy search. We compare it to previously used map-specific adaptive \ac{IPP} formulations for continuous- (\textit{prev-C}) and discrete-valued terrain feature monitoring (\textit{prev-D}) resembling our baselines in \Cref{subsec:exp_setup}. \Cref{T:ablation_study_results} summarises the planning performances .Our map-agnostic adaptive \ac{IPP} formulation consistently performs on par with adaptive \ac{IPP} formulations specifically designed for continuous- and discrete-valued monitoring missions, irrespective of the policy search method. Notably, in some scenarios, our map-agnostic formulation even improves the average planning performance of policy search algorithms. These results verify that our method successfully integrates with state-of-the-art adaptive \ac{IPP} methods without requiring map-specific adaptation. This way, our approach contributes to unifying the broad family of adaptive \ac{IPP} approaches.

%===============================================================================

\section{Conclusion}
\label{sec:conclusion}

We proposed a novel map-agnostic formulation of the adaptive \acf{IPP} problem for terrain monitoring. Our adaptive \ac{IPP} formulation is generally applicable to various continuous- or discrete-valued terrain feature monitoring missions. Our main contribution is a planning state space unifying different map representations. Based on our formulation and a newly introduced reward function, we show how to train a single adaptive \ac{IPP} policy for terrain monitoring missions with varying map representations and user-defined areas of interest. Our experimental results show that our single learned policy performs similarly to state-of-the-art map-specifically designed and trained non-learning- and learning-based adaptive \ac{IPP} methods on simulated and real-world terrain datasets. Our map-agnostic formulation easily integrates with state-of-the-art online policy search methods for adaptive \ac{IPP} while maintaining performance.

%===============================================================================

\bibliographystyle{plain_abbrv}

% All new citations should go to new.bib. The file glorified.bib should go
% be the one from the ipb server. After paper or related work has been
% written merge the entries from new.bib to glorified.bib ON THE SERVER,
% replace the glorified.bib in this repository and empty the new.bib
\bibliography{glorified,new}

\IfFileExists{./certificate/certificate.tex}{
\subfile{./certificate/certificate.tex}
}{}
\end{document}